# Cooperative Pollution Source Localization and Cleanup with a Bio-inspired Swarm Robot Aggregation

Arash S. Amjadi[1], Mohsen Raoufi[2], Ali E. Turgut[3], George Broughton[4], Tomáš Krajník[4] and Farshad Arvin[5]

[1] School of Electrical Engineering, University of Tabriz, Tabriz, Iran. arash.sadeghi@ieee.org
[2] Department of Aerospace Engineering, Sharif University of Technology, Tehran, Iran.
[3] Mechanical Engineering Department, Middle East Technical University, Ankara, Turkey.
[4] Artificial Intelligence Centre, Faculty of Electrical Engineering, Czech Technical University, Prague, Czechia.
[5] Control Systems group, School of Electrical and Electronic Engineering, The University of Manchester, M13 9PL, Manchester, UK. farshad.arvin@manchester.ac.uk

**Abstract.** Using robots for exploration of extreme and hazardous environments has the potential to significantly improve human safety. For example, robotic solutions can be deployed to find the source of a chemical leakage and clean the contaminated area. This paper demonstrates a proof-of-concept bio-inspired exploration method using swarm robotic system, which is based on a combination of two bio-inspired behaviors: aggregation, and pheromone tracking. The main idea of the work presented is to follow pheromone trails to find the source of a chemical leakage and then carry out a decontamination task by aggregating at the critical zone. Using experiments conducted by a simulated model of a Mona robot, we evaluate the effects of population size and robot speed on the ability of the swarm in a decontamination task. The results indicate feasibility of deploying robotic swarms in an exploration and cleaning task in an extreme environment.

## 1 Introduction

Exploration of extreme environments came to the focus of the robotic research community since, as shown by the relatively recent Fukushima disaster, most standard robots fail to operate reliably in extreme environments with chemical and radiation contamination [1]. For instance, in multi-robot systems, wireless communication is widely used to transfer data between robots [2]. However, radiation sources can hamper wireless connections, rendering standard multi-robot teams ineffective. To overcome this problem, one can use short-range communication techniques, common in robotic swarm research, such as an effective inter-robot connection protocol [3], which is resilient to extreme conditions.

In the multi-robot research community, researchers mainly tend to distribute the robot tasks among large numbers of simple robots rather than assigning a complex duty to a single robot because of the numerous advantages of multi-robot systems [4]. Primarily, this is because a group of cooperating robots can carry out a task at a higher speed than when compared to a single robot [5]. A collective decision made by a group of robots will be more reliable than the decision of a single robot in most applications. In addition, multi-robot systems are more fault-tolerant, and improvements in efficiency can be achieved simply by scaling up the number of robots in the swarm. For instance, sensor noise and uncertainties in extreme environments can heavily influence a single robot's decisions. On the other hand, a multi-robot platform can compensate for this uncertainty by processing and combining the information of multiple agents [6]. Moreover, emergent swarm behaviors are also robust to noise and uncertainty of an individual robot's sensors.

Following this trend, robot interaction rules and methods are required in order to establish a protocol by which a number of robots can cooperate with each other and complete a task. In many cases, such protocols contain high level data transfer among robots and decision centers [7]. To establish social behavior among multiple robots, bio-inspired behaviors shown by social animals like ants, bees and termites can be utilized [8]. The observation of effective behaviors in social animals encourages us to choose a deeper and detailed approach to apply a similar scenario for robots. Aggregation is a behavior that is observed in many social animals varying from insects [9] to amoeba [10]. This behavior can be defined as the gathering of individuals around an area with optimal ecologic properties. By forming this aggregation, they gain new abilities like building a habitat, deterring much stronger predators, and defeating larger prey [11]. Aggregation is observed in two types: cue-based aggregation [12], in which nest members aggregate by following an external cue; and self-organized aggregation, where nest members aggregate without any external guidance and regardless of their environment.

For exploration in extreme environments, physical cues (e.g. chemicals or radiation) are crucial, therefore, cue-based aggregation is more relevant. The BEECLUST method, introduced in [13] is one of the most popular bio-inspired aggregation principles. By performing a detailed analysis of honeybees, which follow cue-based aggregation around a zone with a more optimal temperature [14], the BEECLUST aggregation method [13] was chosen as a basis to implement our exploration system. This bio-inspired aggregation was studied intensively in many research works [15], [16], [17].

In terms of communication challenges in swarm robotics, one of the most distinct phenomena of social insects' interactions is their communication method. Social insects often use the environment itself as a communication medium by spreading organic substances (pheromones) that can indicate a multitude of environmental features from the presence of food to dangerous animals. The interaction of pheromones and individual agents lead to efficient swarm behaviors capable of solving chaotic and complex situations. The principles of pheromone communication were utilized by swarm robots in various research studies [18], [19], [20] and its potential for robotic applications has been demonstrated.

In this paper, a method for exploring environments with extreme conditions, e.g. chemical or radiation leakage is developed using robot swarms. The proposed method combines the BEECLUST algorithm and pheromone following behavior. In this paper, the main goal of robots is to detect and clean the source of a chemical leakage, to decontaminate an entire environment. In particular, we investigate the impact of population size, speed of motion, intensity of environmental cues (contamination magnitude) and coherency of the group, on the efficiency of the cleaning operation.

The rest of this paper is organized as follow: in section 2 the proposed exploration method is described. In section 3 the robotic platform and experiments will be demonstrated. In section 4 the results of the experiments will be analyzed and experimented. Finally, in section 5 the paper will be concluded and future work on this exploration method will be explained.

## 2    Localization & Cleanup Method

In the proposed localization method, a combination of cue-based aggregation [13] and pheromone following behavior [21] was developed (shown in Fig. 1). Robots tend to reach the source of a leakage by following the intensity gradient of the cue (contamination magnitude). This chemotaxis behavior has been shown in nature e.g. ants foraging. The main goal is to reach the source and start the cleaning task with presence of other robots. They stop their motion and start the cleaning task when they detect another robot. Therefore, robots are always in one of three stages: 1) follow a chemical gradient, 2) avoid walls and 3) stop and clean. After a robot-robot collision occurs, robots wait for a while depending on the cue intensity that they sense and when the waiting time is over, they continue their initial task i.e. following a chemical gradient. The duration of this cleaning time, $\omega_s$, relies on the intensity of the cue where robots find other robots. Hence, this cleaning time is calculated using the following equation:

$$\omega_s = \omega_{max} \frac{\overline{S_c}}{\overline{S_c}^2 + 25000} , \qquad (1)$$

where $\omega_{max}$ is the maximum waiting time and $\overline{S_c}$ is the average reading from left and right sensors, $\overline{S_c} = \frac{s_r + s_l}{2}$, $0 \leq \overline{S_c} \leq 255$.

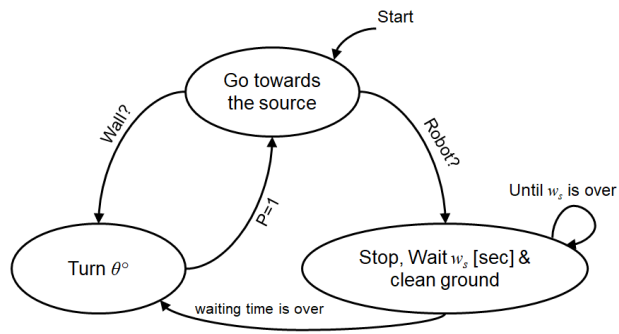

**Fig. 1.** Proposed exploration method. The ovals indicate stages and arrows indicate transitions.

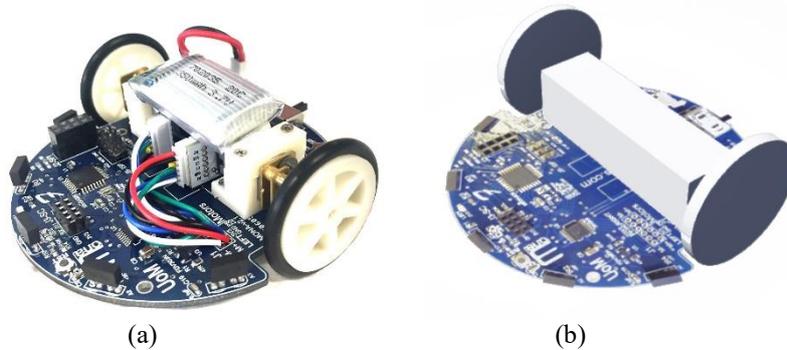

**Fig. 2.** (a) Mona an open-source low-cost robot developed for swarm robotics and (b) Mona model in Webots

When the waiting time is over, robots make a turn of $\theta$ degrees where $\theta$ is a random variable with uniform distribution in the range of [90º 180º] in both the clockwise and counter-clockwise directions. After the random turn, robots continue to follow the chemical gradient in order to reach its center.

## 3 Realization of the Method

In this section, the realization of aforementioned exploration method will be described.

### 3.1 Robotic Platform

The proposed exploration scenario was implemented using Mona robots [22]. It is a miniature wheeled robot with diameter of 8 cm.

**Simulated Model.** Mona was modeled [23] in Webots software to mimic the real Mona robot and proposed method was applied via simulation model of Mona. Fig. 2 shows the model used in Webots for Mona and the real Mona robot. IR proximity sensors are utilized in modeled Mona in order to detect collision. In order to sense the cue, as it was mentioned previously, two sensors beneath right and left wheels of robots are added. In simulation, two sensors facing towards the ground was used and after calibration, the value changes in domain of [0 255] where 0 is a region with no chemical leakage and 255 is the region with the maximum possible intensity of leakage.

**Motion Control.** To describe the algorithm by which robots follow the cue, two sensors are utilized under the left and right wheels of Mona in order to measure the cue intensity under these wheels. The values of these sensors are used in order to determine the desired rotation of Mona in a direction that cue increases so robots will move towards the zone with the highest chemical leakage. To achieve this objective, required value of right wheel, $N_r$, and left wheel, $N_l$, were computed using the equation below:

$$N_r = \frac{s_l - s_r}{\alpha} + \beta \quad \text{and} \quad N_l = \frac{s_r - s_l}{\alpha} + \beta \;, \tag{2}$$

where $s_l$ is the value extracted from the left cue sensor and $s_r$ is the value of the right sensor. The coefficient $\alpha$ adjusts the sensitivity of motors to the difference of sensor values in a way that with lower values of $\alpha$ robots will react dramatically to small differences between two sensors. In this paper, $\alpha$ is set to 2. The last parameter $\beta$ is the biasing coefficient of motors with possible range of [0 10]. For the modeled Mona $\beta= 6$ is equal to speed of 8.0 cm.s$^{-1}$ (speed of one Mona length per second) and consequently, $\beta= 3$ is equal to speed of 4.0 cm.s$^{-1}$ (speed of half Mona length per second).

**Interaction with Environment.** To simulate the environment, a square shaped arena with the length of 285 cm is considered which encompasses a circular shaped white region (contaminated zone) with a diameter of 222.7 cm that decreases linearly as it gets far from the center of cue. Fig. 3 at $t= 0$ s shows the described cue. To demonstrate the cleaning task in a greater detail, the equation below is utilized while robots are in their waiting period:

$$\overline{S_n} = \overline{S_c} - (8 - \sqrt{p^2 + q^2}) \;, \tag{3}$$

where $\overline{S_c}$ is the average value of cue measured by sensors and $\overline{S_n}$ is the value of cue after being cleaned. $p$ and $q$ are variables in range of [-4 4] and they were utilized to give gradient decrease shape to the cleaned area. As it is apparent, the cleaning level will have its maximum value of 8 per second when $p= 0$ and $q= 0$ so the center of squared shaped cleaned region will be darker and the cleaning level will have its minimum value of 2.34 per second when $|p| = |q| = 4$. Therefore, the outer layers will be cleaned less in comparison with center. This equation will be executed once in a second while robot is waiting. Fig. 3 illustrates how a cleaned cue will look like in $t= 1000$ s and $t= 4000$ s.

### 3.2 Experimental Setup

To provide results for exploration method, Webots software is used which is a simulation software for robotic platforms. In Webots software, a supervisor is utilized to set initial conditions of robots and to track simulation to keep record of results. At the beginning of each simulation, robots are located randomly with uniform distribution by the supervisor. In addition, they get random rotation with uniform distribution. The reason that initial state of robots is randomized is to make results independent from the initial positions and rotations of robots.

The effect of two parameters – speed of robot and population of swarm – on the performance of the exploration are investigated. Experiments have conducted in five different population sizes of $N= \{10, 20, 30, 40, 50\}$ robots each one with two different speeds of a robot length per second ($v_r= 8.0$ cm.s$^{-1}$) and a half robot length per second ($v_r= 4.0$ cm.s$^{-1}$). Each set of experiments were repeated six times for the duration of 4000 s. Table I shows the list of parameters and variables that were used in this work.

**Table 1.** Parameters and variables used in paper

| Parameter | Description | Value / range |
|---|---|---|
| $N$ | number of robots | {10, 20, 30, 40, 50} |
| $v_r$ | robot's linear velocity | {4.0, 8.0} cm.s$^{-1}$ |
| $r_c$ | radius of the cue centre | 0.7 m |
| $t$ | time of experiment | [0 4000] s |
| $\omega_s$ | waiting time of robots | [0 21.7] s |
| $\omega_{max}$ | maximum waiting time | 30 s |
| $s_r$ | right sensor value | [0 255] |
| $s_l$ | left sensor value | [0 255] |
| $N_r$ | speed of right wheel | [0 10] |
| $N_l$ | speed of left wheel | [0 10] |

### 3.3 Metric

Three metrics were selected to evaluate the performance of robots: the change of average cue during time, number of robots inside a circular zone with radius of $r_c$ and finally, the coherency of robots. Additionally, in order to analyze the effect of parameters of the system on each of the above-mentioned metrics, we utilized ANOVA (Analysis of Variance) test whose result will be presented subsequently. For each test, we have considered time, population, and the speed of robots as factors. In the following subsections greater details about these three metrics will be provided.

**Average Cue vs Time.** Average cue intensity is measured once in a second, hence by considering the overall experiment duration, 4000 samples of cue level are collected in each test case and the median of measured cue intensities are used for 6 experiments in order to evaluate the swarm performance. The main purpose of this measurement was to find out the impact of change in population size in the performance of robots. Besides population size, the effect of robots' speed in overall performance is also studied.

**Ratio of Robots Close to Cue Center.** The ratio of robots that are within a circle with the same center as cue and with radius of $r_c$= 0.7 m is computed each second for $N$= {10, 30, 50} robots and $v_r$= 8.0 cm.s$^{-1}$ in order to evaluate the performance of robots. Also, the cleaning pattern of robots can be recognized by this metric.

**Coherency.** A metric by which the coherency of the swarm is evaluated during the run time is coherency distance. The variable is calculated by averaging the distance of all robots to each other.

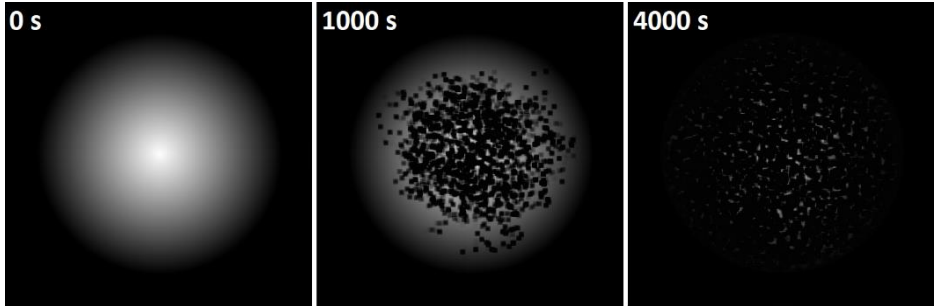

**Fig. 3.** Evolution of cue during 4000 s with $N= 30$ robots.

## 4 Results & Discussion

To provide an example of how a chemical leakage disappears, Fig. 3 shows the chemical leakage states at $t= \{0, 1000, 4000\}$ s for $N= 30$ robots with $v_r= 8.0$ cm.s$^{-1}$. At $t= 4000$ s, it could be said that cue has been disappeared.

### 4.1 Average Cue vs Time

Fig. 4 demonstrates results of two experiment cases. In each case, experiments were conducted with five different populations where each population was repeated 6 times and the median of 6 experiments was plotted. At first glance, it could be deduced that according to the results, robots with $v_r= 8.0$ cm.s$^{-1}$ could decrease the intensity of chemical leakage faster in comparison with robots that move by velocity of $v_r= 4.0$ cm.s$^{-1}$. It is also apparent that as the number of robots increases, so does the vanishing speed of chemical leakage. Therefore, similar to the results of aggregation experiments in [15], the growth of population significantly improved swarm performance. It can be seen from Fig. 4 that the best performance among test cases belongs to $N= 50$ robots with $v_r= 8.0$ cm.s$^{-1}$ where the weakest performance belongs to $N= 10$ robots with $v_r= 4.0$ cm.s$^{-1}$. In addition, as it was proposed in [24], it is observed that test cases with high population achieve their objective of cleaning chemical leakage in shorter time which proofs the scalability of the swarm system. The results were also statistically analyzed to find the most effective factor. Table II shows the results of ANOVA test, that reveal all parameters (time, population size, and speed of robot) significantly affect the system ($p \leq 0.05$), however, speed of robot was the most significant factor ($F= 67.144$). On the hand, the least effective factor was time ($F= 5.032$) that shows the cleaning process is less time-dependent.

### 4.2 Ratio of Robots Close to Cue Center

The ratio of robots that are close to source of leakage with a distance of less than 0.7 m for $N= \{10, 30, 40\}$ robots and $v_r= 8.0$ cm.s$^{-1}$ to all robots of that experiment case were plotted in Fig. 5. This ratio decreases during time and remains constant at about $t=1500$ s for $N=\{30, 50\}$ robots and for $N=10$ robots it can be observed that ratio of robots close to the center of cue does not change considerably. The time $t=1500$ s (the time that ratio remains unchanged) is observed in Fig. 4 for $N=\{30, 50\}$ robots and $v_r=8.0$ cm.s$^{-1}$. It can be understood that for both populations the intensity of cue has been reduced more than 50% by that time. So as proposed, there is an obvious connection between this metric and previous one since when cue almost disappears, robots do not aggregate close to the center anymore and they move to places with higher intensity of cue.

**Table 2.** Results of ANOVA test for the cue intensity

| Factors | p-value | F-value |
| --- | --- | --- |
| Time | 0.000 | 5.032 |
| Population | 0.000 | 39.383 |
| Speed | 0.000 | 67.144 |

As proposed in metrics section, the cleaning pattern of robots can be determined from Fig. 5 since it can be seen that the ratio of robots which are close to the center of cue (with distance less than $r_c=0.7$ m) increases dramatically while cue has not been decreased by 50%. After the noticeable increase of the ratio, it decreases to reach its steady state. From this manner of ratio, it can be concluded that at first thousand seconds, most of the robots are within the radius of $r_c=0.7$ m and they distribute from center as the leakage vanishes. Therefore, robots clean the cue from inside, starting by the highest leakage points and continuing to points with less leakage. When the cue almost disappears, they do their random walk and at that moment, measurements in all presented plots remain steady.

To compare the impact of population on ratio of robots close to the cue center, it can be observed that as the population increases, the steady state ratio decreases since the center of cue has vanished and robots tend to spread all over the environment randomly. However, steady state ratio will be higher for smaller population since there is still some visible parts of the cue. Beside comparing the steady state ratio, the slope of plots before reaching steady state also heavily depends on population, such that higher population will have sharper decreases before their steady state since they will clean the cue quicker and more.

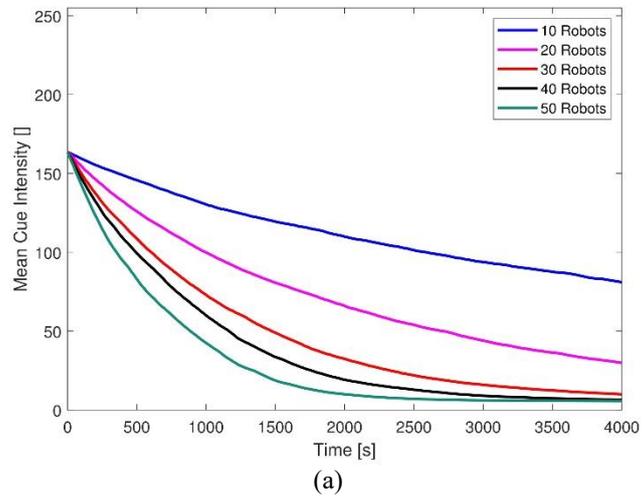

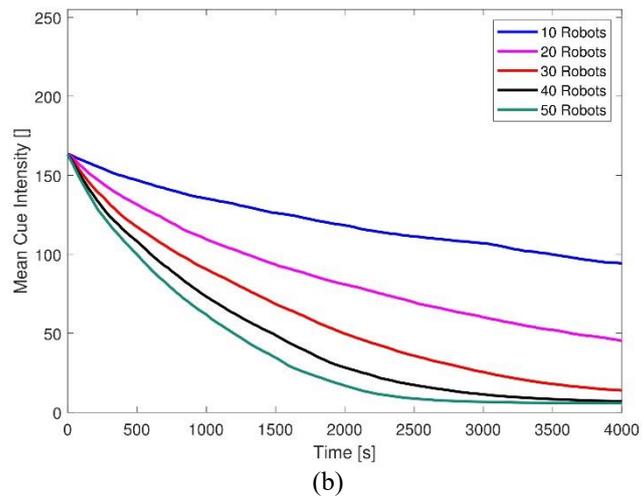

**Fig. 4.** Mean intensity of the cue vs time for various swarm population and robot speeds. (a) $v_r$= 8.0 cm.s$^{-1}$ (b) $v_r$= 4.0 cm.s$^{-1}$

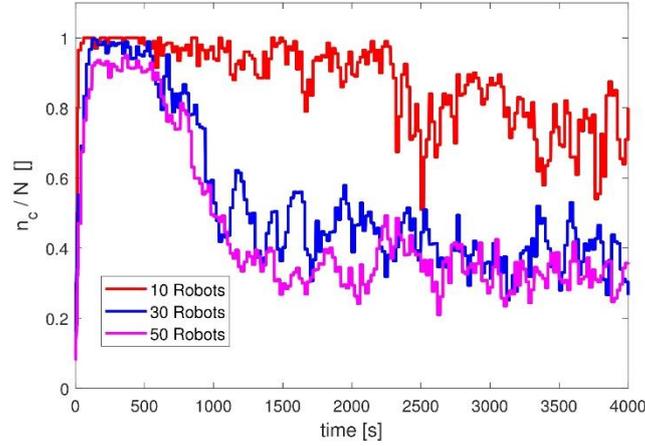

**Fig. 5.** Ratio of robots within a distance of $r_c$= 0.7 m from the source of leakage for $N$= {10, 30, 50} robots with speed of $v_r$= 8.0 cm.s$^{-1}$

### 4.3 Coherency

Fig. 6 demonstrates the coherency of three population sizes of $N$= {10, 30, 50} robots with $v_r$= 8.0 cm.s$^{-1}$. The similar point of test cases is that they verge to a steady state after a while. On the other hand, they differ in the slope of verging to this steady value as for higher population sizes, coherency reaches its steady state faster. In order to define steady state for coherency, we shall consider $N$= 50 robots and compare its results with previous metrics. In Fig. 4, it can be seen that for $N$= 50 robots with $v_r$= 8.0 cm.s$^{-1}$ speed, after $t$= 2500 s the average cue intensity does not change considerably. If the same moment is considered in Fig. 6, it can be seen that at that moment coherency also does not change considerably. Therefore, it can be deduced that whenever chemical leakage vanishes, coherency and average cue reach their steady state. There are similar features between Fig. 5 and Fig. 6. Primarily, in both figures for $N$= 10 robots, plots do not change considerably and as the population grows, plots in both figures fluctuate more. For $N$= {30, 50} robots, plots in both figures remain unchanged after $t$= 1500 s. To analyze this behavior, when coherency remains unchanged, it means that there is no longer a cooperation among robots. At that moment, ratio of robots close to the center of source also remains constant so the robots are not following cue any more hence they mainly tend to perform random walk. Consequently, from these observations we can conclude that the more the cue disappears, tendency of robots for random walk increases and their cooperation decreases.

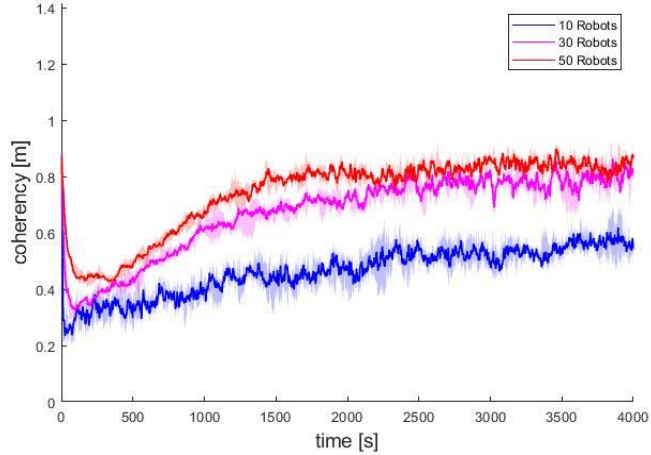

**Fig. 6.** Median coherency measured in meters vs time for $N = \{10, 30, 50\}$ robots with $v_r = 8.0$ cm.s$^{-1}$. Shades around the plots indicates the maximum and minimum coherency.

To compare the effect of population size in coherency, it can be seen that as the population grows, the coherency of the robots changes more dramatically. It can be observed that for $N = 10$ robots coherency does not alter noticeably. In contrast, for $N = 50$ robots, coherency encounters a dramatic decrease meaning that the robots get very close to each other while cue is not cleaned. The results from this section were also statistically analyzed. These results showed that all the factors significantly affected the coherency of the swarm system ($p \leq 0.05$), however, population was the most significant factor ($F = 19.088$) in coherency of the robots' motion.

**Table 3.** Results of ANOVA test for the Coherency

| Factors | p-value | F-value |
| --- | --- | --- |
| Time | 0.000 | 5.214 |
| Population | 0.000 | 19.088 |
| Speed | 0.000 | 5.305 |

Although the main goal of this work was to demonstrate an application of swarm robotics in an extreme environment, such system will also be useful in other applications e.g. agri-robotics [25], industrial machinery [26] and odor localization [27].

## 5   Conclusion & Future Work

In this paper, a chemical leakage localization and cleanup scenario was implemented using robotic swarm. The method is based on a bio-inspired aggregation scenario for swarm robots. Impacts of population and speed on swarm performance were studied.

We also studied the evolution of cooperation between robots and the ratio of robots present near the source of leakage in order to track the robots' interaction during process and as the results showed, robots cooperation decreases as the leakage vanishes. Therefore, these results can be significant guide for developing real world application of swarm robotics in an extreme environment. In future works, we will consider multiple leakage sources with dynamic behavior. We will use artificial neural networks for robots in order to increase their performance for use in dynamic environments.